\documentclass[10pt,twocolumn,letterpaper]{article}

\usepackage{cvpr}

\definecolor{cvprblue}{rgb}{0.21,0.49,0.74}
\usepackage[pagebackref,breaklinks,colorlinks,allcolors=cvprblue]{hyperref}
\usepackage{multirow}
\usepackage{times}
\usepackage{epsfig}
\usepackage{graphicx}
\usepackage{amsmath}
\usepackage{amssymb}

\usepackage{booktabs}
\usepackage{epigraph}
\usepackage{enumitem}
\usepackage{float}
\usepackage{chngcntr}
\usepackage{soul}
\usepackage{caption}
\usepackage{subcaption}
\usepackage{setspace} 
\usepackage{zref-xr}
\usepackage{colortbl}
\usepackage{capt-of} %
\usepackage{etoolbox} %
\usepackage[accsupp]{axessibility}
\definecolor{baselinecolor}{gray}{.8}

\newcommand{\ours}{MotiF\xspace} %
\newcommand{\benchmark}{TI2V-Bench\xspace}

\def\ie{\emph{i.e}\onedot} 
 
 \def\vs{\emph{vs}\onedot}

\def\etal{\emph{et al}\onedot}

\def\paperID{***}
\def\confName{CVPR}
\def\confYear{2025}

\title{MotiF: Making Text Count in Image Animation with Motion Focal Loss}

\author{
Shijie Wang$^{1}$\footnotemark[2]
\qquad
Samaneh Azadi$^{2}$
\qquad
Rohit Girdhar$^{2}$
\qquad
Saketh Rambhatla$^{2}$ \\
Chen Sun$^{1}$
\qquad
Xi Yin$^{2}$
 \\
$^{1}$ Brown University \quad $^{2}$ GenAI, Meta \\
{\tt\small \{shijie\_wang, chensun\}@brown.edu, \{azadis, rgirdhar, rssaketh, yinxi\}@meta.com}}

\begin{document}
\twocolumn[{%
\renewcommand\twocolumn[1][]{#1}%
\maketitle
\begin{center}
    \centering
    \captionsetup{type=figure}
    \includegraphics[,clip,width=0.95\linewidth]{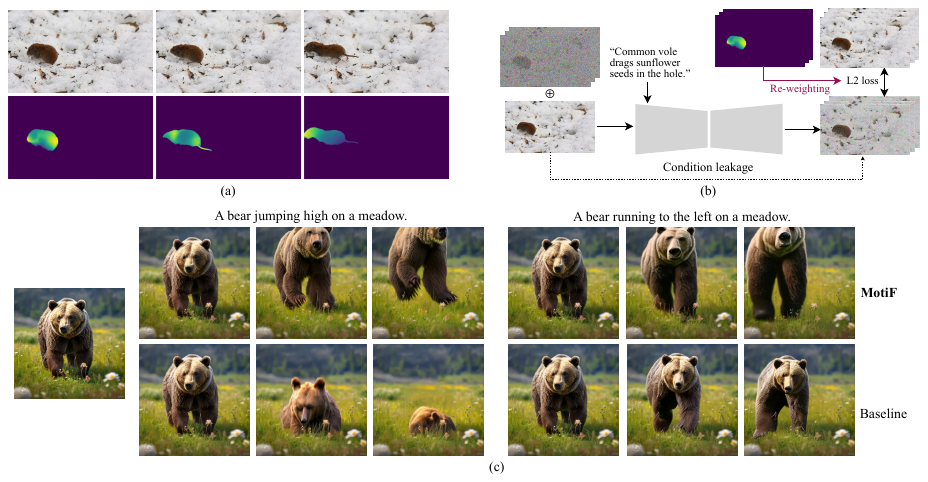}
    \captionof{figure}{\textbf{Motivation and results of \ours.} (a) Example video frames and the corresponding motion heatmaps calculated from optical flow. In this example, $97\%$ of the pixels are static while only $3\%$ has meaningful motion. (b) In standard TI2V training pipeline, the model may learn to over-rely on the conditional image to optimize the L2 loss. This issue has been identified in~\cite{zhao2024identifying} and termed as conditional image leakage. We propose \ours to guide the model's learning to focus on regions with more motion via motion heatmap re-weighting. (c) Qualitative results comparing \ours to the baseline on examples from our proposed \benchmark evaluation set.}
    \label{fig:teaser}
\end{center}%
}]

\renewcommand{\thefootnote}{\fnsymbol{footnote}}
\footnotetext[2]{This work is done during Shijie's internship at Meta.}
\renewcommand{\thefootnote}{\arabic{footnote}}

\begin{abstract}
Text-Image-to-Video (TI2V) generation aims to generate a video from an image following a text description, which is also referred to as text-guided image animation. Most existing methods struggle to generate videos that align well with the text prompts, particularly when motion is specified. To overcome this limitation, we introduce \ours, a simple yet effective approach that directs the model's learning to the regions with more motion, thereby improving the text alignment and motion generation. We use optical flow to generate a motion heatmap and weight the loss according to the intensity of the motion. This modified objective leads to noticeable improvements and complements existing methods that utilize motion priors as model inputs. 
Additionally, due to the lack of a diverse benchmark for evaluating TI2V generation, we propose \benchmark, a dataset consists of $320$ image-text pairs for robust evaluation. We present a human evaluation protocol that asks the annotators to select an overall preference between two videos followed by their justifications. Through a comprehensive evaluation, \ours outperforms nine open-sourced models, achieving an average preference of $72\%$. The \benchmark and additional results are released in the \href{https://wang-sj16.github.io/motif/}{project website}.
\end{abstract}
    
\section{Introduction}
\label{sec:intro}
From creating short clips of moving MNIST digits~\cite{mittal2017sync} to producing a minute-long high-definition video featuring a stylish walking woman~\cite{videoworldsimulators2024}, Text-to-Video (T2V) generation~\cite{ho2022video, singer2022make, blattmann2023stable, polyak2024movie} has progressed remarkably over the past decade. 
Meanwhile, video generation with additional conditions~\cite{li2023videogen, qing2024hierarchical, girdhar2023emu, wang2024microcinema, wu2023tune,wei2024dreamvideo} has also gained considerable attention.
In this paper, we focus on Text-Image-to-Video (TI2V) generation, a task first introduced in~\cite{Hu_2022_CVPR} aiming to generate a video based on an initial image and a text description~\cite{xing2025dynamicrafter, zhang2023i2vgen, zhao2024identifying, ma2024follow}. 
Our goal is to improve text adherence in open-domain TI2V generation.
Commonly known as Image-to-Video (I2V) generation ~\cite{zhang2023i2vgen, zhao2024identifying} or image animation~\cite{xing2025dynamicrafter,ma2024cinemo}, this task often overlooks the text guidance.
In many cases, text merely describes the image or video~\cite{huang2024vbench,fan2024aigcbench}, or is unnecessary~\cite{yu2024zero, liu2025physgen, chen2024animating}, underscoring the need for focused improvement.

The main challenge in TI2V modeling is that, while the image offers a strong spatial signal to learn the video content, the model must rely mainly on the text to infer motion. 
When given an image and multiple different prompts, most previous work generates videos with limited motion. Zhao \etal~\cite{zhao2024identifying} identified this issue as conditional image leakage, where the model overly relies on the image condition.
To address this issue, some studies propose to reduce the strength of the image condition by adding noise or masking the image~\cite{zhao2024identifying, ma2024follow}.
Others seek to derive more motion priors to facilitate motion learning~\cite{chen2025livephoto, zhang2024pia}.
All of these methods are based on enhancing the input signals, in the hope that the model can learn to leverage them implicitly.
In contrast, we propose to improve the training objective to enable the model to focus on the regions with more motion explicitly, which is an orthogonal direction to prior works and can be easily combined with existing techniques. 

We hypothesize that the model's difficulty in following motion-based instructions may stem from insufficient attention to motion patterns.
As shown in Figure~\ref{fig:teaser} (a), in a video with a static background, $97\%$ of the pixels remain unchanged over time, while only $3\%$ exhibit meaningful motion. 
This subtle motion pattern might be neglected by the model especially considering that all regions are treated equally in the L2 loss.
Motivated by this, we introduce {\bf Moti}on {\bf F}ocal loss (\ours), a term inspired by the focal loss in dense object detection~\cite{ross2017focal}.
Specifically, we first use optical flow~\cite{teed2020raft} to create a motion heatmap that represents the motion intensities of the video.
We then use the motion heatmap to assign loss weight for the video to focus on regions with more motion. 
Additionally, we analyze commonly used image conditioning mechanisms and observe that concatenating the conditioning image with the diffusion input works best for both image and text alignment.

Beyond training, evaluating TI2V generation is also challenging due to the lack of a suitable evaluation set and reliable metrics.
Existing evaluation sets are limited, they are either too small~\cite{zhang2024pia} or the text is used as a description of the starting image rather than specifying the intended motion for animation~\cite{fan2024aigcbench}.
Therefore, we introduce a new benchmark,~\benchmark, 
that consists of $320$ image-text pairs covering $22$ different scenarios. Each scenario includes $3$ to $5$ images with similar content in different styles, and $3$ to $5$ prompts to animate these images to generate videos with different motion. We include challenging scenarios such as novel object introductions and fine-grained object references. We use the Emu model~\cite{dai2023emu}, publicly available through \url{meta.ai}, to generate the initial images.

Previous studies have used various automatic metrics to assess different quality aspects, such as video quality, motion intensity, and text alignment. We observed that these metrics may not always align with human perception.
Following Movie Gen~\cite{polyak2024movie}, we mainly rely on human evaluation with A-B testing to evaluate the performance.
Our human evaluation protocol is motivated by the JUICE protocol~\cite{girdhar2023emu}, in which annotators are asked to indicate an overall preference and justify their choices across several axes, including image alignment, text alignment, object motion, and overall quality. 
Through our comprehensive human evaluations, we show that \ours improves TI2V generation capabilities, notably improving text alignment and motion quality.
In summary, we make the following contributions.
\begin{itemize}
\item We present Motion Focal Loss (\ours) that encourages TI2V generation to concentrate on regions with larger motion and is complementary to existing techniques. 
\item We present a new benchmark, \benchmark, including synthetic images and texts that cover a wide range of complex scenarios, as well as a human evaluation protocol designed to assess the performance of TI2V generation.
\item Through comprehensive comparisons with nine previous methods, we demonstrate the effectiveness of \ours, achieving an average preference of $72\%$.
\end{itemize}

\section{Related Work}
\label{sec:related}

\subsection{Text-Image-to-Video Generation}
With the rapid advancements in Text-to-Image (T2I) and Text-to-Video (T2V) generation, TI2V generation has also gained significant attention.
Some training-free methods~\cite{yu2024zero, ni2024ti2v} utilize pretrained T2I or T2V models to enable zero-shot TI2V generation. Our review is focused on training-based approaches, 
~\cite{an2023latent,zhou2022magicvideo,singer2022make}. 
The techniques in prior TI2V generation methods can be mainly grouped into two categories: 1) design novel architectures to integrate the image condition, 2) derive motion priors to improve motion learning.
Our approach falls into the second category.

\noindent\textbf{Image Condition Integration.}
Most recent works use either the U-Net architecture~\cite{ho2022video, ho2020denoising} or the Diffusion Transformer (DiT)~\cite{peebles2023scalable} for video generation, where the image condition is incorporated similarly. 
Most works concatenate the image with the noisy video~\cite{girdhar2023emu, blattmann2023stable, ma2024cinemo, chen2025livephoto}, which acts as a spatial-aligned content guidance.
Additionally, DynamiCrafter~\cite{xing2025dynamicrafter}, ConsistI2V~\cite{ren2024consisti2v} and LivePhoto~\cite{chen2025livephoto} also use cross-attention layers on the image embeddings to further \textit{strengthen} the image condition and improve consistency in video generation. On the contrary, Zhao~\etal~\cite{zhao2024identifying} apply higher noise levels at later time steps to \textit{weaken} the image condition to alleviate the condition image leakage issue. 
Follow-Your-Click~\cite{ma2024follow} introduce a masking strategy to degrade the image condition. 
In \ours, instead of disrupting the image condition to shift the model's focus on the text, our training objective explicitly encourages the model to focus on video regions with more motion, and thus be more faithful to the text description.

\noindent\textbf{Motion Priors.}
Our world exhibits a rich diversity of unique motion patterns. Modeling motion from real-world videos for generation is challenging due to the inherent ambiguity that arises when a single description can correspond to multiple possible motions.
Many methods explore motion priors to improve motion learning. 
A common approach is to calculate a motion score~\cite{chen2025livephoto, dai2023animateanything, ma2024follow, zhang2024pia, qing2024hierarchical, ma2024cinemo, guo2024i2v} from the video and embed it into the model, which also enables motion intensity control during inference. 
LivePhoto~\cite{chen2025livephoto} introduce a framework to re-weight text embeddings to place greater emphasis on the motion-related words.
Cinemo~\cite{ma2024cinemo} use a novel strategy to learn motion residuals. 
 MoCA\cite{yan2023motion} leverage optical flow as an additional conditioning.
Similar to our work, Follow-Your-Click~\cite{ma2024follow} also use optical flow to calculate a mask, but the mask is embedded into the model and it requires additional input from the user during inference time. 
In our case, the motion prior is leveraged at the training objective level, which is explicit, simple and does not require any additional input during inference.

\subsection{Video Generation Evaluation}
\noindent\textbf{Benchmarks and Automatic Metrics.}
Make-a-Video reports FVD/IS on the UCF-101 dataset~\cite{soomro2012ucf101} and FID/CLIP similarities on the MSR-VTT dataset~\cite{xu2016msr} under the zero-shot setting, which is followed by later works in both T2V~\cite{blattmann2023align, zhou2022magicvideo, an2023latent, ge2023preserve} and TI2V generation~\cite{xing2025dynamicrafter, ren2024consisti2v, ma2024cinemo, zhao2024identifying}. 
Re-purposing existing video understanding datasets for TI2V evaluation is undesirable due to their unconstrained data collection procedures, which may not capture the nuances of TI2V tasks.
As a result, recent studies start to collect new evaluation sets~\cite{ren2024consisti2v, fan2024aigcbench, zhang2024pia} and explore additional automatic metrics~\cite{liu2024evalcrafter, huang2024vbench, zhang2024benchmarking}. 
AIGCBench~\cite{fan2024aigcbench} is proposed for TI2V task. 
However, their text descriptions are used to generate the image but not to drive the motion generation.
AnimateBench~\cite{zhang2024pia} shares some similarities with our benchmark as using T2I models to generate images and applying different text prompts to animate the same image.
However, the number of unique images and texts in AnimateBench is small and the domains are also limited. 

\noindent\textbf{Human Evaluation.}
Following Movie Gen~\cite{polyak2024movie}, we rely on human annotators to evaluate the performance of TI2V generation.
Prior works use different human evaluation protocols including A-B testing~\cite{zhao2024identifying}, A-testing~\cite{chen2025livephoto} and ranking multiple methods~\cite{ma2024follow, guo2024i2v, zhang2024pia}. 
While A-testing and ranking are more efficient, we believe A-B testing is less ambiguous and may yield more reliable results.
Unlike previous works that evaluate multiple metrics, we ask the annotators to select the best video and justify their choices based on various aspects, following the JUICE design~\cite{girdhar2023emu}. 
This will generate a single metric for calibration while also enabling a detailed examination of results through the various quality justifications.
To the best of our knowledge, we have conducted the most comprehensive human evaluation by comparing to nine open-sourced TI2V models and showing the advantage of \ours.

\section{Approach}
\label{sec:method}

\begin{figure*}[t]
    \centering
    \includegraphics[width=0.85\linewidth]{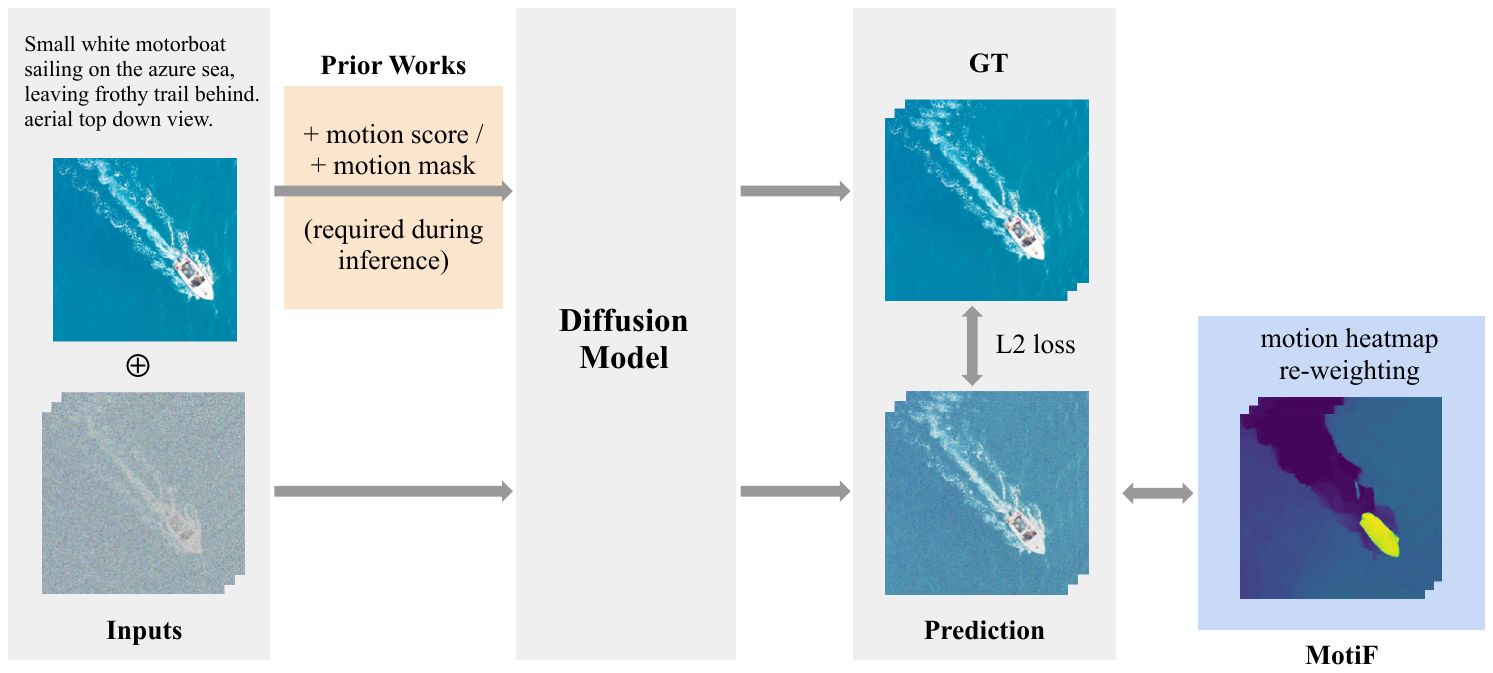}
    \captionof{figure}{\textbf{High-level comparisons of \ours \vs prior works.} Previous TI2V generation methods mainly focused on deriving additional motion signals (motion score and/or motion mask) as inputs for the model to leverage implicitly. On the contrary, we focus on the learning objective and propose to weight the diffusion loss based on the motion intensity, that is derived from optical flow. Our method is simple, effective, and does not require additional inputs during inference. Moreover, \ours is complementary to existing techniques.}
    \label{fig:overview}
\end{figure*}

Given an image $x_i$ and a text prompt \( \mathbf{c} \), TI2V task aims to generate an $L$-frame video \( \mathbf{x} = \{x_0, x_1, \dots, x_L\} \) that follows the text description. In this work, we specifically tackle a common scenario when the image $x_i$ is the first frame (\ie $i = 0$).
The generated video should maintain visual coherence with the starting image and produce motion that is driven by the text description.
In this section, we present the details of our model architecture, the proposed motion focal loss, and the procedure of generating the motion heatmaps.

\subsection{Preliminaries}
\noindent \textbf{Video Diffusion Models.}
Diffusion models~\cite{sohl2015deep, ho2020denoising, song2020score} are probabilistic generative models designed to create images and videos from random Gaussian noise. In the forward pass, a diffusion process gradually transforms a data sample \( \mathbf{x}_0 \sim p_{\text{data}}(\mathbf{x}) \) into Gaussian noise \( \mathbf{x}_T \sim \mathcal{N}(\mathbf{0}, \mathbf{I}) \) over \( T \) timesteps by sequentially adding noise according to \( q(\mathbf{x}_{t}|\mathbf{x}_{t-1}) \). 
The diffusion model learns the reverse denoising process, where it iteratively removes noise to reconstruct the data, predicting a less noisy \( \mathbf{x}_{t-1} \) from \( \mathbf{x}_t \).

An input video with \( L \) frames is represented as a 4D tensor \( \mathbf{x} \in \mathbb{R}^{L \times H \times W \times 3} \). This high-dimensional data introduces considerable computational demands, especially for high-resolution or long-duration videos. To address this challenge, latent video diffusion models (LVDMs)~\cite{an2023latent,zhou2022magicvideo} operating within the latent space are commonly used in video generation to reduce the computation cost. 
Specifically, frames are encoded into the latent space via \( \mathbf{z} = \mathcal{E}(\mathbf{x}) \), where \( \mathcal{E}: \mathbb{R}^{L \times H \times W \times 3} \rightarrow \mathbb{R}^{L' \times H' \times W' \times C'} \) denotes the encoder. The diffusion process \( \mathbf{z}_t = q(\mathbf{z}_{t-1}, t) \) and denoising process \( \mathbf{z}_{t-1} = p_{\theta}(\mathbf{z}_{t}, \mathbf{c}, t) \) are then conducted in the latent space, where \( \mathbf{c} \) represents the input conditions. Finally, the decoder decodes the generated video \( \hat{\mathbf{x}} = \mathcal{D}(\hat{\mathbf{z}}_0) \) from the denoised latent prediction \( \hat{\mathbf{z}}_0 \). 
Most prior works use the image-based Variational Auto-Encoder~\cite{rombach2022high}  to encode each frame independently so that $L'=L$.

During training, the input of the denoising model is a linear combination of the video latent and Gaussian noise based on the time step. 
The objective is to optimize the model to predict the noise (or other variants), typically with mean squared error:

\begin{equation}
    \begin{aligned}
\mathcal{L}_{\text{diffusion}} = \mathbb{E}_{t, \mathbf{x} \sim p_{\text{data}}, \mathbf{\epsilon} \sim \mathcal{N}(\mathbf{0}, \mathbf{I})}\Vert\mathbf{\epsilon}-\mathbf{\epsilon}_{\theta}\left(\mathbf{z}_t, \mathbf{c} ,t\right)\Vert_2^2.
    \end{aligned}
\end{equation}

\subsection{Modeling}
\label{sec:model}
As shown in Figure~\ref{fig:overview}, different from prior works that all focused on incorporating more motion priors into the model, and thus may require additional inputs during inference, we focus on improving the training objective to explicitly focus on the motion learning during training.  

\noindent \textbf{Motion Focal Loss.} 
The diffusion loss treats all the latents equally across the spatial and temporal dimensions. 
However, the motion is not evenly distributed across the video. 
It's common that most of the video stays static while only a small region has meaningful motion. 
As a result, successive frames tend to closely resemble the initial frame, meaning that a ``still video'', \ie duplicating the initial frame, can yield a relatively low loss. 
To address this, we introduce the motion focal loss, \( \mathcal{L}_{\text{motif}} \), to explicitly focus on high-motion regions during training. 

We begin first by generating a motion heatmap \( \mathbf{m} \in \mathbb{R}^{L \times H \times W} \) for each video, where each entry \( \mathbf{m}[l, h, w] \in [0, 1] \) represents the motion intensity at position \([h, w]\) in the \( l \)-th frame of the input video \( \mathbf{x} \).  
The motion heatmap \( \mathbf{m} \) is down-sampled to \( \mathbf{m}' \in \mathbb{R}^{L \times H' \times W'} \) to align with video latents \( \mathbf{z} \). The motion focal loss is defined as:

\begin{equation}
    \begin{aligned}
\mathcal{L}_{\text{motif}} = \mathbb{E}_{t, \mathbf{x} \sim p_{\text{data}}, \mathbf{\epsilon} \sim \mathcal{N}(\mathbf{0}, \mathbf{I})}\,  \Vert \mathbf{\mathbf{m}'} \cdot \left(\epsilon - \mathbf{\epsilon}_{\theta}\left(\mathbf{z}_t, \mathbf{c}, t \right) \right ) \Vert_2^2.
    \end{aligned}
\end{equation}

The model is trained with a joint loss with \( \lambda \) scaling  the motion focal loss relative to the diffusion loss:
\begin{equation}
    \begin{aligned}
\mathcal{L} = \mathcal{L}_{\text{diffusion}} + \lambda \mathcal{L}_{\text{motif}}.
    \end{aligned}
\end{equation}

\noindent \textbf{Motion Heatmaps.}
The use of motion heatmaps is to enhance the model's ability to generate motion by concentrating on areas with significant activity. Although there may be several methods to achieve this, we first explore using optical flow to create these motion heatmaps.
To generate the motion heatmap \( \mathbf{m}_l \) for the \( l \)-th frame \( \mathbf{x}_l \), we first compute the optical flow intensity \( \mathbf{f}_l \) between \( \mathbf{x}_l \) and the subsequent frame \( \mathbf{x}_{l+1} \) using RAFT~\cite{teed2020raft}. We then apply a sigmoid-like function \( \mathbf{f}_l \) to normalize the intensity map to be within the range \([0, 1]\), enhancing contrast in the motion distribution and yielding the final motion heatmap where higher value denotes large motion.

\begin{figure*}[ht]
    \centering
    \includegraphics[width=\textwidth]{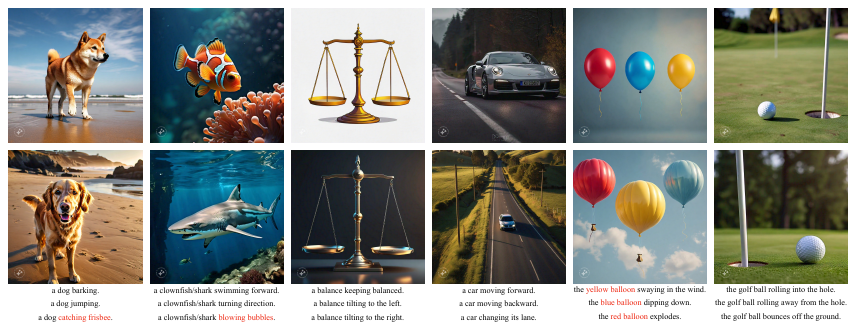}
    \caption{\textbf{Example image-text pairs in \benchmark}. For each scenario (column), we first think of a scene that could be potentially animated to generate different types of motion. We include challenging scenarios when there are multiple objects ({\textcolor{red}{yellow/blue/red balloon}}) in the initial image for fine-grained control or the text prompt describes a new object ({\textcolor{red}{frisbee, bubbles}}) to enter the scene. Then we come up with different prompts and use the publicly available \url{meta.ai} tool to generate diverse sets of images. Images of low quality or those not in the appropriate initial state are removed.}
    \label{fig:benchmark}
\end{figure*}

\begin{table*}[!t]
    \centering
    \scalebox{0.75}{
    \begin{tabular}{cccccccc}
        \toprule
         Name & Type & Text-Driven & Media-Text Pairs & Unique Media & Unique Text & Evaluation & Metrics \\
         \midrule
         I2V-Bench~\cite{ren2024consisti2v} & video-text  & No & $2,950$ & $2,950$ & $2,950$ & auto & visual quality, visual consistency \\
         AIGCBench~\cite{fan2024aigcbench} & video-text & No & $1,000$ & $1,000$ & $1,000$ & auto & MSE, SSIM, CLIP, etc \\
         AIGCBench~\cite{fan2024aigcbench} & image-text & No & $925$ & $925$ & $925$ & auto & MSE, SSIM, CLIP, etc \\
         AIGCBench~\cite{fan2024aigcbench} & image(synthetic)-text & No & $2,003$ & $2,003$ & - & auto & MSE, SSIM, CLIP, etc \\ 
         Animate Bench~\cite{zhang2024pia} & image(synthetic)-text & Yes & $105$ & $35$ & $~16$ & auto & image alignment, text alignment\\ \hline 
         \benchmark (Ours) & image(synthetic)-text & Yes & $320$ & $88$ & $133$ & human & TI2V score \\
         \bottomrule
    \end{tabular}
    }
    \caption{\textbf{Recent TI2V evaluation benchmarks.} We believe the key for TI2V generation is that the text should describe the motion, termed as text-driven in the table. While Animate Bench is similar to our benchmark, it's very specific for the personalization domain and the dataset size and diversity are limited in both the images and the text descriptions. Moreover, the proposed metrics are all image-level evaluations that do not account for motion.} 
    \label{tab:benchmark}
\end{table*}

\noindent \textbf{Image Conditioning.}
We build our model based on a pre-trained T2V diffusion model, VideoCrafter2~\cite{chen2024videocrafter2}, to speed up the training process.
We inject the image condition to the T2V model for TI2V generation. 
Previous work such as DynamiCrafter~\cite{xing2025dynamicrafter} perform a dual-stream image injection, \ie feed the visual features using the cross-attention layers (referred as {\textit{cx-attn}}) and concatenating the image latent with the noised video (referred as {\textit{x-cat}}) to improve the consistency in video generation.

We observe that using cx-attn for image conditioning presents two main drawbacks. Firstly, when cx-attn is used alone, it may generate videos with poor content and style alignment with the input image. Secondly, when combined with x-cat, it may compete with the text features in the cross-attention layers, reducing the model's ability to follow text prompts effectively.
Therefore, we inject the image condition by solely concatenating the image latent with the noised video latents, \ie, x-cat. 
Following~\cite{chen2024videocrafter2}, we use an FPS embedding module to embed the FPS signal similar to the timestep embedding, and CLIP text encoder for cross-attention text condition.

\section{\benchmark}
\label{sec:benchmark}

\subsection{Evaluation Set}
While it's relatively easy to collect a diverse set of prompts for T2V evaluation, evaluating TI2V is not straightforward as it requires a conditional image. 
As shown in Table~\ref{tab:benchmark}, recent TI2V benchmarks can be classified into three categories: 1) video-text pairs; 2) image-text pairs with realistic images; 3) image-text pairs with synthetic images. 
Re-purposing existing video understanding datasets~\cite{soomro2012ucf101,xu2016msr,bain2021frozen} for TI2V evaluation is non ideal, as these datasets assume only one possible outcome from the initial frame, and the text descriptions typically do not emphasize the intended motion.
Existing images are also limited in their diversity and potential motion changes. 
Thus, we resort to image-text pairs with synthetic images to curate \benchmark. 
AIGCBench~\cite{fan2024aigcbench} and Animate Bench~\cite{zhang2024pia} are most relevant to our work. 
However, the texts in AIGCBench are used to generate the images but not meant to describe the motion in the videos, which is critical for TI2V task. 
Animate Bench employs motion-based text descriptions, but its evaluation set is relatively small and limited, where the images often feature a single object or scene, and the text description lacks diversity.

Our goal is to curate an evaluation set that has a diverse set of images, and each image has a diverse set of prompts for text-guided animation. 
Some examples are shown in Figure~\ref{fig:benchmark}.
To create this benchmark, we first design scenarios in which multiple actions can be performed from an initial state and enumerate a range of possible actions. 
For example, in the scenario of ``an aerial view of a car on the road," possible actions could include ``moving forward," ``moving backward," and ``changing its lane."
We come up with $22$ different scenarios with diverse objects and scenes, each with multiple options for different motion. 
We make sure to include challenging scenarios like images with multiple objects, and text descriptions that introduce a new object expected to appear in the video.

For each scenario, we aim to generate $3$ to $5$ images with varying appearance and styles to ensure dataset diversity. 
We design text prompts to generate multiple starting images and iteratively refine the prompts to improve the image quality. 
We use the publicly available \url{meta.ai} to generate all the images. 
Low quality examples or those unable to support all the predefined actions are filtered out. 
In the example above, images were excluded if the vehicle's front direction was unclear or if lane lines were not distinctly visible. 
Our final evaluation set consists of $320$ image-text pairs including $88$ unique images and $133$ unique prompts.  
To the best of our knowledge, this is the most comprehensive and challenging benchmark to evaluate TI2V generation.

\begin{figure*}[t]
    \centering
    \includegraphics[width=\textwidth]{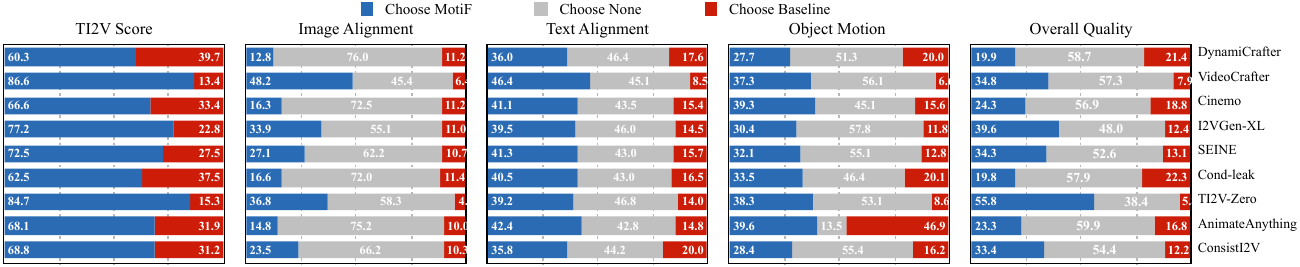}
    \caption{\textbf{Human evaluation results comparing \ours to nine open-sourced models~\cite{xing2025dynamicrafter, chen2023videocrafter1, ma2024cinemo, zhang2023i2vgen, chen2023seine, zhao2024identifying, ni2024ti2v, dai2023animateanything, ren2024consisti2v} on \benchmark}. We achieved considerable improvements across the board with an average preference of $72\%$. Through examining the justification choices, we found that our model mostly excel at improving text alignment and object motion, which matches very well with our motivation.}
    \label{fig:main_result}
\end{figure*}

\subsection{Human Evaluation}
Most T2V and TI2V works conduct human evaluation to assess the overall quality and the text alignment of the generated videos using various metrics. 
The issue with evaluating separate, independent questions is that mixed results make it challenging to reach a clear conclusion.
Therefore, we aim to have a single metric to evaluate the performance while allow more detailed examination of different quality aspects for further analysis, following the JUICE protocol~\cite{girdhar2023emu}. 

Specifically, each A-B comparison consists of two questions. 
We intentionally do not have the equal option so annotators are forced to choose the best one and justify their choices. 
First, annotators are asked to indicate their overall preference between two videos in the context of text-guided image animation.
Second, they are required to justify their choice based on four different aspects: 1) object motion (not only camera movement); 2) alignment with the text prompt; 3) alignment with the starting image; 4) overall quality. 
Annotators can select any combinations of these aspects. 
The overall preference is used to indicate the model's performance in TI2V generation, which we refer to as TI2V score. 
This evaluation protocol allows us to generate a single metric to draw conclusions for A-B comparisons while enabling a more detailed analysis.

\section{Experiment}
\label{sec:experiment}

\begin{figure*}[t]
    \centering
    \includegraphics[width=\linewidth]{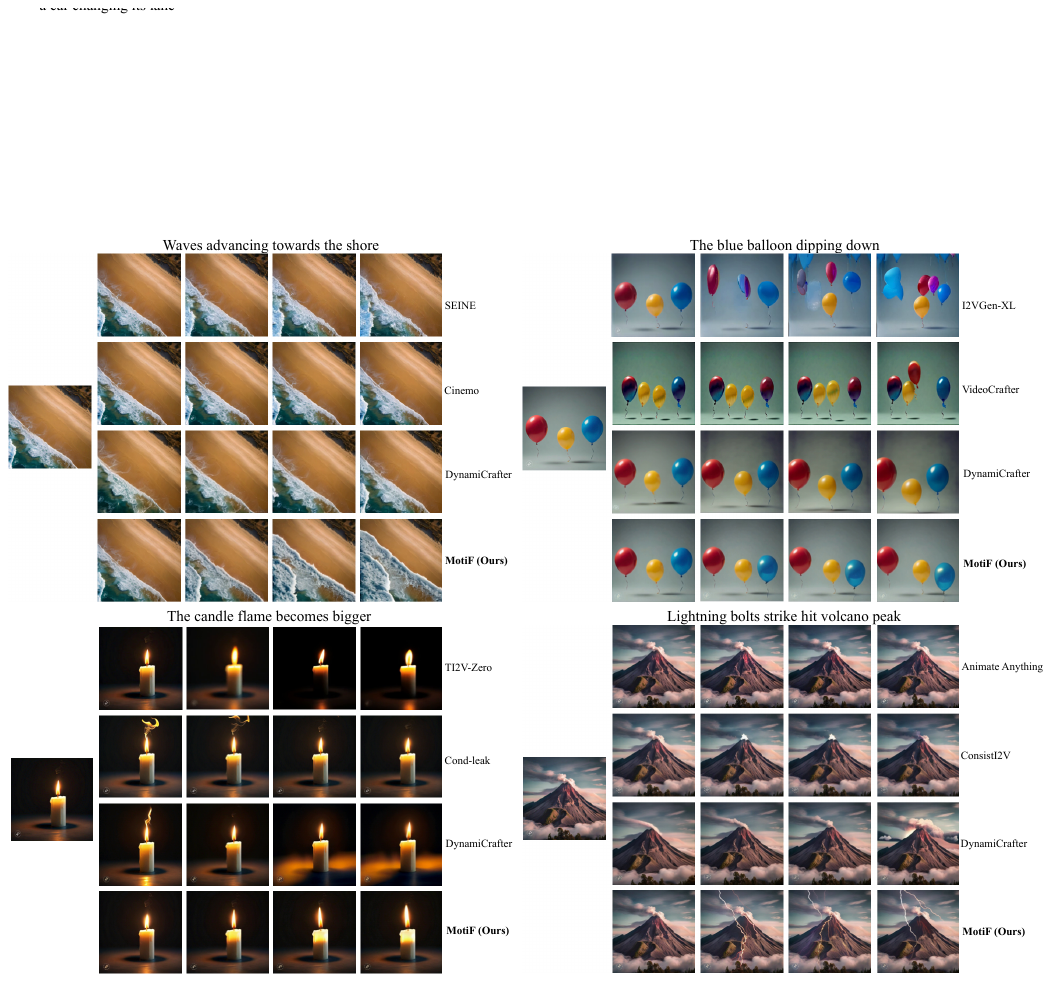}
    \captionof{figure}{\textbf{Qualitative comparison to prior works on \benchmark.} Sampled frames are ordered from left to right.}
    \label{fig:comparison}
\end{figure*}

\subsection{Implementation Details}
\noindent \textbf{Model Architecture.} 
We build our model based on a pretrained T2V model, VideoCrafter2~\cite{chen2024videocrafter2} (@$512$ resolution). 
This model uses CLIP as the text embedding, a Variational Auto-Encoder (VAE) as the video encoder and decoder, and a spacetime-factorized U-Net as the denoising model. 
We concatenate the condition image with the noised video and expand the first embedding layer, similar to prior works. 

\noindent \textbf{Training and Inference.} 
We use an internal licensed dataset of $1$M video-text pairs that is similar to~\cite{xing2025dynamicrafter} for training. 
Videos are center-cropped and sampled at \(320 \times 512\) resolution of $16$ frames with a dynamic frame stride ranging from $1$ to $6$ following~\cite{xing2025dynamicrafter}.
The model is optimized in the v-prediction mode~\cite{salimans2022progressive} by a combination of the diffusion loss and the motion focal loss with (\(\lambda = 1\)). 
The optical flow intensity is normalized using a sigmoid-like function \( \sigma(x) = 1 / (1 + e^{100(0.05-x)}) \), which generates a continuous and polarized heatmap within the range of \([0, 1]\).
The text is randomly dropped by $10\%$ to enable classifier-free guidance. 
We use a constant learning rate of \(5 \times 10^{-5}\), a global batch size of $64$, and a linear noise schedule with $1000$ diffusion steps. 
We train the model for $32$K steps on $8$ A100-80G GPUs. 
During inference, we use the DDIM sampler~\cite{song2021denoising} with $50$ steps and a guidance scale of $7.5$.

\noindent \textbf{Human Evaluation.}
We conduct human evaluations using Amazon Mechanical Turk. 
Annotators are instructed to make their selections only after watching both videos, with a minimum submission time requirement of one minute to ensure thorough assessments. 
We random shuffle the order of the videos shown to annotators to avoid bias. 
For each comparison, we ask $5$ annotators to evaluate the results and perform majority vote.%
We use the overall preference as the TI2V score.
We also calculate a score for each justification criterion as the percentage of times a particular aspect is chosen for the model in the entire evaluation set without majority vote.

\subsection{Comparison to Prior Works}
We compare to nine open-sourced TI2V generation methods. \footnote{Note that all the inference of prior works are done at Brown University.}
We ensure fair comparisons by following each method's pre-processing and post-processing pipeline to ensure the input is in the optimal aspect ratio and the generated videos are without any undesirable distortion.
The results are shown in Figure~\ref{fig:main_result}.
Our method wins by a considerable margin to prior works. 
From the justification selections, the main reasons for our method to win are on the object motion and text alignment, which is exactly the motivation of \ours to improve text-driven motion learning.

Figure~\ref{fig:comparison} shows the qualitative results where \ours can generate higher quality videos that align better with the text prompts, even on challenging cases with multiple objects in the scene with fine-grained motion animation. 
I2VGen-XL and VideoCrafter generate videos that do not align well with the condition image because they use only cross-attention for the image condition embedding, which does not have the spatial alignment as in concatenation. 

In addition to human evaluation, we also report the automatic evaluation on Animate Bench~\cite{zhang2024pia} that is most relevant to \benchmark. As shown in Table~\ref{tab:auto_eval}, \ours achieved comparable performance to prior works. It's worth noting that a static video baseline (by repeating the condition image) gets the best Image Alignment score and reasonable Text Alignment score, suggesting that these metrics are non ideal for evaluating TI2V generation. This has motivated us to mainly rely on human evaluation.

\begin{table}[t]
    \centering
    \scalebox{0.65}{
    \begin{tabular}{cccccc}
        \toprule
         \multirow{2}{*}{Loss} & TI2V & Image & Text & Object & Overall \\
         & Score  & Alignment & Alignment & Motion & Quality  \\
         \midrule
         w/o \ours loss & ${\bf{63.1}}/36.9$ & $10.3/{\bf{10.7}}$ & ${\bf{34.9}}/16.4$ & ${\bf{32.9}}/16.4$ & $18.2/{\bf{20.8}}$ \\
         w/ Inv-\ours loss & ${\bf{61.9}}/38.1$  & $7.6/{\bf{9.2}}$  & ${\bf{34.8}}/12.8$  & ${\bf{34.9}}/15.4$  & $14.3/{\bf{17.8}}$ \\
        
         \bottomrule
    \end{tabular}
    }
    \caption{\textbf{Ablation studies on different design choices}. The numbers on the left is for \ours and the right is for the baseline. Similarly to the comparisons to prior works, \ours mostly excel in improving the text alignment and object motion.}
    \label{tab:ablation_loss}
\end{table}

\begin{table}[t]
    \centering
    \scalebox{0.68}{
    \begin{tabular}{cccccc}
        \toprule
         \multirow{2}{*}{Image Condition} & TI2V & Image & Text & Object & Overall \\
         & Score  & Alignment & Alignment & Motion & Quality  \\
         \midrule
         cx-attn + x-cat & ${\bf{58.1}}/41.9$ & ${\bf{15.0}}/14.6$ & ${\bf{31.5}}/21.7$ & ${\bf{34.0}}/21.6$ & $18.8/{\bf{19.3}}$ \\
         cx-attn & ${\bf{92.2}}/7.8$ & ${\bf{56.8}}/5.3$ & ${\bf{33.8}}/3.8$ & ${\bf{41.3}}/4.5$ & ${\bf{28.2}}/5.7$ \\
         \bottomrule
    \end{tabular}
    }
    \caption{\textbf{Ablation study on the image conditioning methods}. Compared to our choice ({\textit{x-cat}}), {\textit{cx-attn} alone leads to much worse results and using both is also sub-optimal}. Here we train the models without motion focal loss to simplify the setting.}
    \label{tab:ablation_image_cond}
\end{table}

\begin{figure}[t]
    \centering
    \includegraphics[trim={0 0 0 0},clip,width=0.32\textwidth]{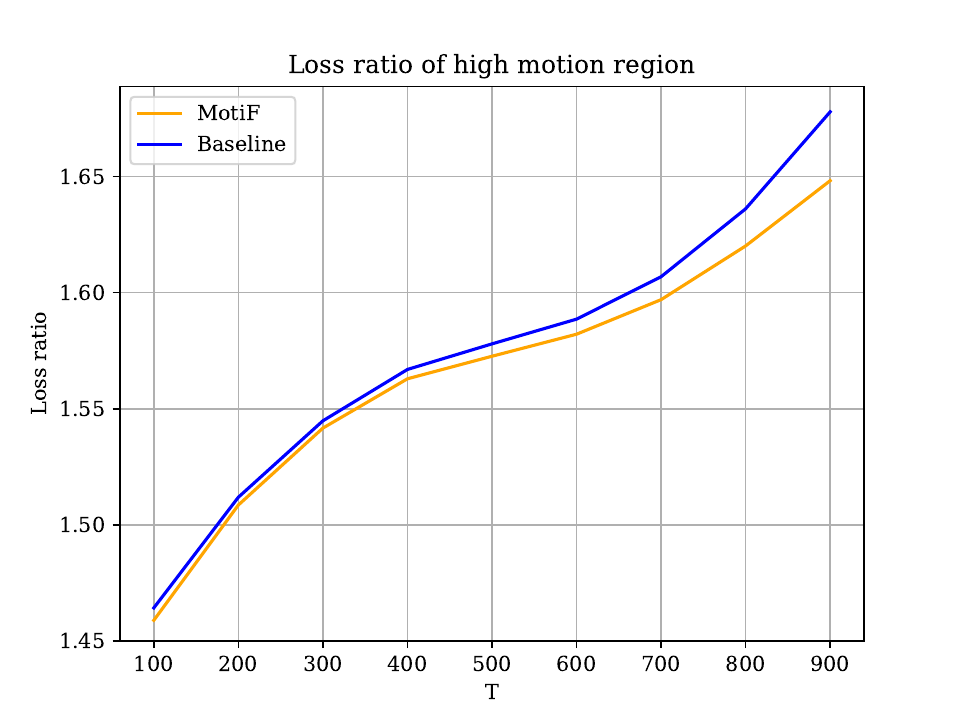
}
    \caption{\textbf{Loss comparison.} We calculate the ratio of the average loss in the high motion region to the average overall loss on a hold-out validation set with different timesteps. \ours can effectively reduce the relative loss of the high motion regions.}
    \label{fig:loss}
\end{figure}

\begin{table}[h]
    \centering
    \scalebox{0.95}{
    \begin{tabular}{ccc}
        \toprule
        \multirow{2}{*}{Method} & Image & Text \\
           & Alignment $\uparrow$ & Alignment $\uparrow$ \\ \hline
        Baseline: static & $99.29$ & $66.24$ \\ \hline
        Cinemo~\cite{ma2024cinemo} & $93.28$ & $66.16$  \\
        Cond-leak~\cite{zhao2024identifying} &$94.05$ &$66.56$\\
        DynamiCrafter~\cite{xing2025dynamicrafter} & $93.41$ & $66.58$ \\ 
        AnimateAnything~\cite{dai2023animateanything} &$96.78$ &$66.64$ \\
        VideoCrafter~\cite{chen2024videocrafter2} & $84.45$ & $66.94$\\
        SEINE~\cite{chen2023seine} & $91.55$ & $67.22$ \\ 
        I2VGen-XL~\cite{zhang2023i2vgen} & $87.13$ & $67.97$ \\ 
        TI2V-Zero~\cite{ni2024ti2v} & $73.78$ & $68.89$ \\ 
        ConsistI2V~\cite{ren2024consisti2v} &$91.33$ &$67.38$ \\ \hline
        \ours (ours) & $92.68$ & $67.73$ \\
         \bottomrule
    \end{tabular}
    }
    \caption{\textbf{Automatic metrics on Animate Bench~\cite{zhang2024pia}.} A simple static video baseline (repeating the first frame) can generate the best image alignment score and reasonable text alignment score (first row). \ours achieved comparable results to prior works.} 
    \label{tab:auto_eval}
\end{table}

\subsection{Ablation Studies}
\noindent \textbf{Motion Focal Loss.}
To verify the effectiveness of \ours, we compare to a baseline model trained without \( \mathcal{L}_{\text{motif}} \) while keeping all other training settings constant. 
As shown in the first row of Table~\ref{tab:ablation_loss}, \ours results in a considerable gain compared to the baseline. 
We also experiment with a Inv-motion loss, where the inverse of the motion heatmap is applied when calculating \( \mathcal{L}_{\text{motif}} \), causing the model to focus on modeling the static regions instead. 
As shown in the second row of Table~\ref{tab:ablation_loss}, \ours achieves much better scores, further validating the effectiveness of explicitly modeling high-motion areas for TI2V generation.

To understand how \ours helps to improve motion generation, we calculate the diffusion loss on a hold out validation set of $2000$ samples for both the baseline and \ours. For each sample, we apply a threshold on the motion heatmap to get a binary motion mask. We then calculate the loss ratio of the high motion region, which is defined as the average diffusion loss of the mask region divided by the average loss overall. As shown in Figure~\ref{fig:loss}, \ours achieves smaller loss ratio compared to the baseline across different diffusion time-steps, again, validating the effectiveness of \ours in improving motion generation.

\noindent \textbf{Image Conditioning.}
As discussed in Section~\ref{sec:method}, one key design space for TI2V generation is on how to integrate the image condition. 
We use concatenation ({\textit{x-cat}}) and ablate the other two choices: 1) cross-attention ({\textit{cx-attn}}); 2) {\textit{x-cat}} and {\textit{cx-attn}}.
As shown in Table~\ref{tab:ablation_image_cond}, using {\textit{cx-attn}} alone results in much worse performance in all metrics. This is understandable as the spatial information is missing.
However, using {\textit{cx-attn}} together with {\textit{x-cat}} also leads to obvious degradation especially on text alignment and object motion, we hypothesis that this is due to the competition of text and image embedding in the cross-attention layers.

\section{Conclusions}
\label{sec:conclusion}
In this paper, we focus on the often overlooked problem of text alignment in text-guided image animation. We hypothesis that the reason prior works struggle to follow the text prompts is because the model may not pay attention to the motion patterns during training. Thus, we present \ours to guide the model's learning on regions with more motion. In addition, we curate a challenging benchmark to evaluate TI2V generation. Although \ours have shown considerable advantages over prior works, it is still limited in generating high quality videos with coherence motion in challenging scenarios when there are multiple objects or new object is expected to enter the scene. We hope this work will attract more attention in solving this challenging problem.

{
    \small
    \bibliographystyle{ieeenat_fullname}
    \bibliography{main}
}

\clearpage
\setcounter{table}{0}
\renewcommand{\thetable}{A\arabic{table}}
\setcounter{figure}{0}
\renewcommand{\thefigure}{A\arabic{figure}}

\appendix

\section*{Appendix}
\label{sec:appendix}

This appendix includes the following sections:
\begin{itemize}
\item Additional Ablation Study (Sec.~\ref{sec:supp_ablation}): 
\item Additional Qualitative Results (Sec.~\ref{sec:supp_viz})
\item Additional Implementation Details (Sec.~\ref{sec:supp_details})
\end{itemize}

\section{Additional Ablation Study}
\label{sec:supp_ablation}
\subsection{Motion Heatmap}
\label{sec:supp_heatmap}
Beyond using optical flow to generate the motion heatmap, we explore alternative approaches leveraging the Segment Anything Model 2 (SAM 2)~\cite{ravi2024sam2}, a state-of-the-art video segmentation model. SAM 2 produces object masks by accepting points or bounding boxes as prompts, generating an initial mask for the first frame, and propagating it consistently across subsequent frames. To target objects with significant motion, we compute optical flow intensity in the first frame and sample \( K \) high-intensity points as prompts for SAM 2. To ensure spatial diversity, the frame is divided into an \( M \times N \) grid, and the top-\( K \) grids with the highest average intensity are selected. From each grid, the point with the highest intensity is chosen as the prompt. SAM 2 then generates masks for the corresponding objects across frames, producing a binary motion heatmap. In our experiments, we set \( K = 5 \) and use a \( 5 \times 5 \) grid.

As shown in the last row of Table~\ref{supp_tab:ablation_heatmap}, the continuous heatmap generated via optical flow achieves better results than the binary SAM-based heatmap especially on text alignment and object motion while SAM-based heatmap improves the overall quality. We observe that although SAM 2 can generate high-quality instance masks across frames with appropriate prompting, it is less effective than optical flow at highlighting large motion regions. We believe that improving prompt sampling strategies could further enhance the performance of SAM 2-based heatmap.

\begin{table}[h]
    \centering
    \scalebox{0.8}{
    \begin{tabular}{ccccc}
        \toprule
          TI2V & Image & Text & Object & Overall \\
         Score $\uparrow$ & Alignment $\uparrow$ & Alignment $\uparrow$ & Motion $\uparrow$ & Quality $\uparrow$ \\
         \midrule
         ${\bf{58.8}}/41.3$ & $11.4/{\bf{12.1}}$ & ${\bf{34.0}}/21.6$ & ${\bf{31.2}}/22.4$ & $15.0/{\bf{20.6}}$ \\
         \bottomrule
    \end{tabular}
    }
    \caption{\textbf{Ablation study comparing SAM heatmap to optical flow heatmap (\ours).} Numbers on the left are for \ours and right for the SAM heatmap.}
    \label{supp_tab:ablation_heatmap}
\end{table}

\subsection{Motion Focal Loss Weight}
\label{sec:supp_loss}
We investigate the impact of the motion focal loss weight \( \lambda \) on \benchmark. As shown in Table~\ref{supp_tab:ablation_scale}, simply setting \( \lambda = 1 \) achieves the best overall results. Generally, reducing \( \lambda \) can improve the overall visual quality of the generated videos but results in lower TI2V Score specially on worse text alignment and object motion.

\begin{table}[t]
    \centering
    \scalebox{0.75}{
    \begin{tabular}{cccccc}
        \toprule
         \multirow{2}{*}{$\lambda$} & TI2V & Image & Text & Object & Overall \\
         & Score $\uparrow$  & Alignment $\uparrow$ & Alignment $\uparrow$ & Motion $\uparrow$ & Quality $\uparrow$ \\
         \midrule
         0.5 & ${\bf{66.6}}/33.4$ & $10.0/{\bf{10.8}}$ & ${\bf{37.3}}/13.7$ & ${\bf{39.8}}/15.3$ & $19.4/{\bf{19.7}}$ \\
         2 & ${\bf{63.8}}/36.3$ & ${\bf{12.4}}/7.6$ & ${\bf{33.0}}/20.4$ & ${\bf{31.7}}/21.1$ & ${\bf{27.5}}/16.1$ \\
         5 & ${\bf{65.6}}/34.4$ & ${\bf{15.2}}/12.8$ & ${\bf{39.7}}/15.9$ & ${\bf{36.8}}/16.5$ & ${\bf{23.1}}/17.8$ \\

         \bottomrule
    \end{tabular}
    }
    \caption{\textbf{Ablation studies on the motion focal loss weight $\lambda$}. The numbers on the left is for \ours and the right is for the comparing setting.}
    \label{supp_tab:ablation_scale}
\end{table}

\section{Additional Quantitative Evaluation}
\label{sec:supp_auto}
In the paper, we mainly rely on human evaluation for comparisons. 
Here, we provide additional automatic evaluation results for baseline models on VBench-I2V~\cite{huang2023vbench}.

\begin{table*}[t]
    \centering
    \scalebox{0.8}{
    \begin{tabular}{cccccccccccc}
        \toprule
        \multirow{2}{*}{Method} & Subject & Background & Temporal & Motion & Dynamic & Aesthetic & Image & I2V & I2V & Camera \\
        & Consistency$\uparrow$ & Consistency$\uparrow$ & Flickering $\uparrow$ & Smoothness$\uparrow$ & Degree $\uparrow$ & Quality$\uparrow$ & Quality$\uparrow$ & Subject$\uparrow$ & Background$\uparrow$ & Motion $\uparrow$ \\
         \midrule
   Baseline: static & 100.00 & 100.0 & 100.0 & 99.84 & 0 & 65.54 & 71.61 & 98.77 & 97.24 & 14.29  \\
      DynamiCrafter & 94.70 & 97.55 & 95.17 & 97.39 & 39.51 & 60.40 & 68.16 & 96.89 & 96.68 & 30.88 \\
      Cinemo        & 96.80 & 99.04 & 98.67 & 98.95 & 17.32 & 59.92 & 64.37 & 97.43 & 98.14 & 15.83 \\ \hline
      MotiF (ours)         & 95.27 & 98.37 & 97.27 & 98.16 & 30.98 & 58.70 & 66.95 & 96.89 & 97.00 & 24.35 \\
         \bottomrule
    \end{tabular}
    }
    \caption{\textbf{Results on VBench-I2V}.}
    \label{supp_tab:vbench}
\end{table*}

\subsection{Evaluation on VBench-I2V}
VBench-I2V~\cite{huang2024vbench} is another popular image-to-video (I2V) benchmark, consisting of 356 real-world images and 1,118 image-prompt pairs. Different from \benchmark, VBench-I2V uses image captions as text conditions instead of action instructions and especially focus on controlling camera motion through text prompts. We evaluate MotiF alongside two strongest baseline models, DynamiCrafter and Cinemo, as well as the static video baseline, on VBench-I2V. 
Results are presented in Table~\ref{supp_tab:vbench}.

MotiF achieves comparable performance to DynamiCrafter and Cinemo in consistency, temporal flickering, motion smoothness, and video quality, while the static video baseline significantly outperforms all models in most metrics except dynamic degree and camera motion. This underscores a key limitation of automatic evaluation: the trade-off between video dynamics and overall quality makes it challenging to provide a holistic assessment of model performance. Additionally, we observe that existing metrics for video dynamics, often based on optical flow, tend to favor videos with significant camera or background motion over object motion. This highlights the importance of conducting human evaluations for the TI2V task to address these shortcomings in automatic evaluation.

\section{Additional Visualization}
\label{sec:supp_viz}

\begin{figure*}[t]
    \centering
    \includegraphics[width=\linewidth]{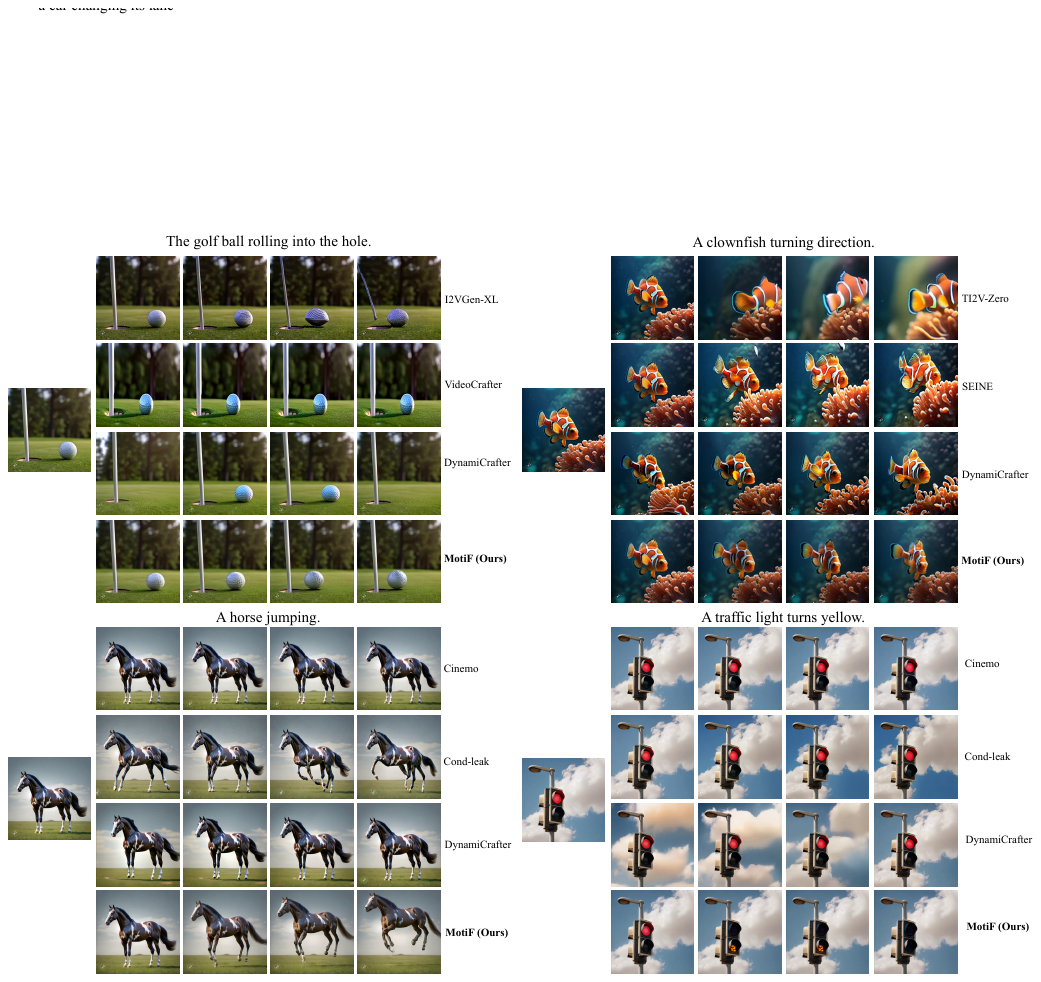}
    \captionof{figure}{\textbf{More qualitative comparison to prior works on \benchmark.} \ours can generate videos that align better with the text prompts. %
    More video samples are available in the project website.}
    \label{fig:comparison_supp}
\end{figure*}

\begin{figure*}[t]
    \centering
    \includegraphics[trim={0 345 0 0},clip, width=0.8\linewidth]{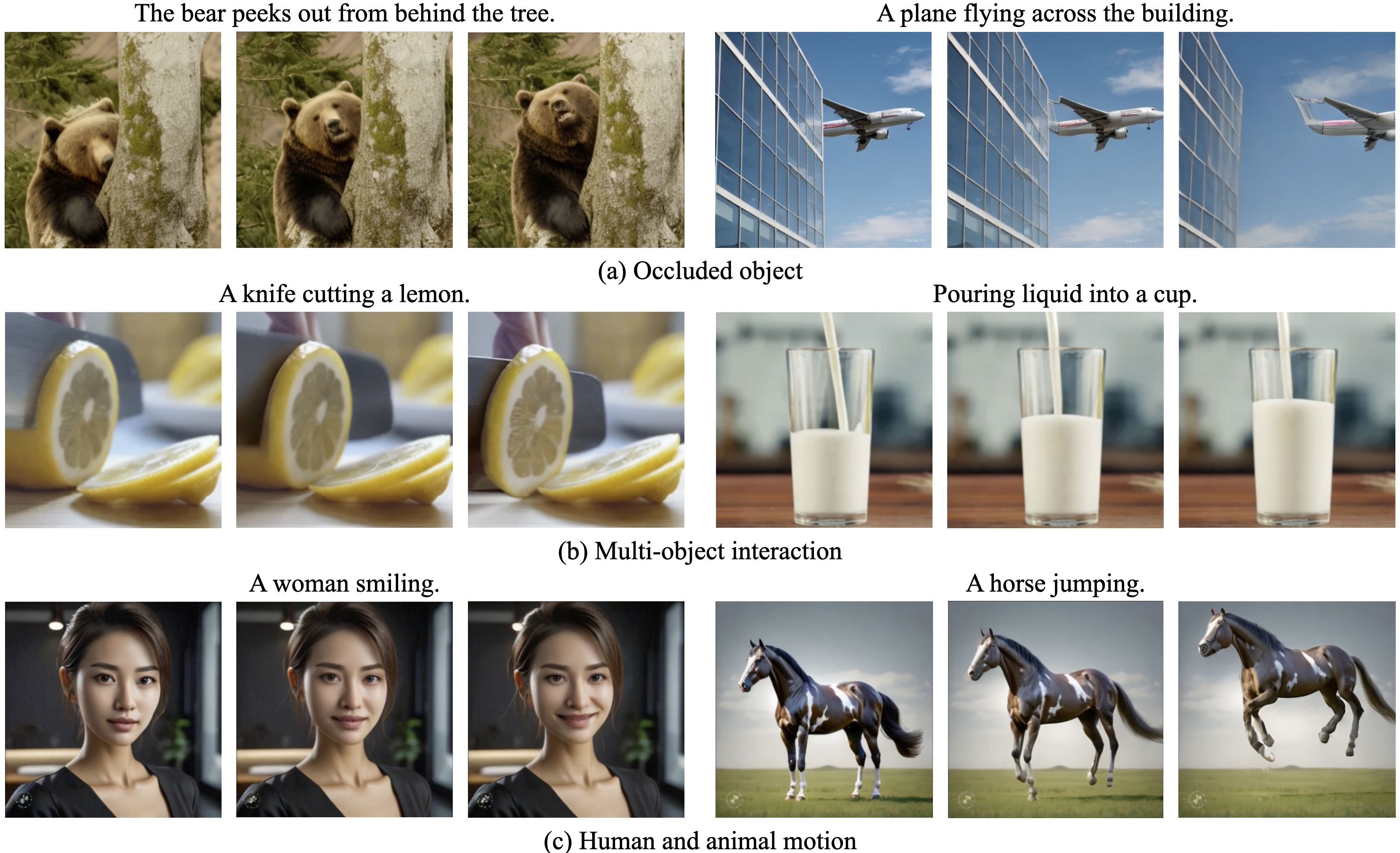}
    \captionof{figure}{\textbf{Results on complex scenarios.} \ours generates faithful videos for (1) object occlusion and (b) multiple object interaction.}
    \label{fig:complex_supp}
\end{figure*}

\begin{figure*}[t]
    \centering
    \includegraphics[width=\linewidth]{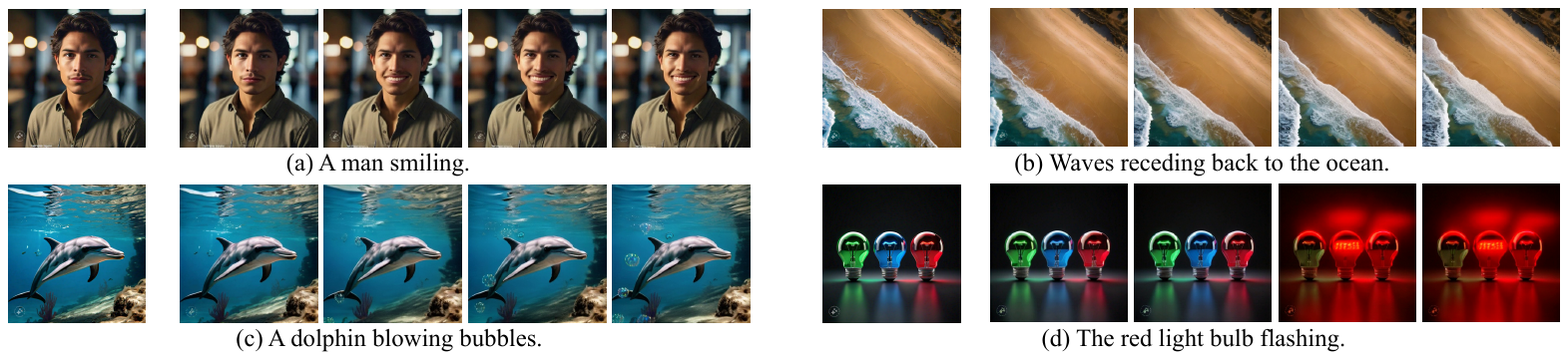}
    \captionof{figure}{\textbf{Typical failure and challenging cases of MotiF on \benchmark.} We observe two typical cases that the model fail: 1) the generated videos may have unnatural motion ((a)); 2) the generated videos do not align well with the prompts ((b), (c), (d)). For 2), there are two specific scenarios when following the text is challenging including novel object ((c)) or multiple objects ((d)). %
   We also include more video samples in the project website.
    }

    \label{fig:failure_supp}
\end{figure*}

\subsection{Comparison with Baseline Models}
Figure~\ref{fig:comparison_supp} shows more qualitative results comparing to prior methods. 
\ours can generate videos that better align with the input text prompts, which validates the effectiveness of the proposed motion focal loss. 
We also include the video samples in the supplementary folder. 

\subsection{Additional Examples}
We also provide additional examples that goes beyond TI2V-Bench for complex scenarios including occlusions and multi-object interaction. The results are shown in Figure~\ref{fig:complex_supp}.

\subsection{Failing Cases Analysis}
Although \ours shows clear advantages over prior work, it is still far from perfect for TI2V generation especially on our proposed challenging benchmark \benchmark. 
As shown in Figure~\ref{fig:failure_supp}, we observe two main types of failure cases. 
First, sometimes the generated motion is not very natural. 
Second, the generated video may not follow the text prompt. 
In the second case, there are two challenging scenarios of \benchmark: 1) when the text prompt describes a new object that needs to appear in the scene, the generated video may not be very coherent; 2) when there are multiple objects and the text prompt only refers to one of them, it will be hard for the model to generate precise motion.

We hope \ours and \benchmark will help the research community to tackle this challenging problem. 
\ours is generic and complementary to existing techniques for TI2V generation. 
We believe that improving the motion heatmap accuracy can potentially boost the performance.
Moreover, \ours is potentially applicable to text-to-video generation.

\begin{figure*}[t]
    \centering
    \includegraphics[width=0.8\linewidth]{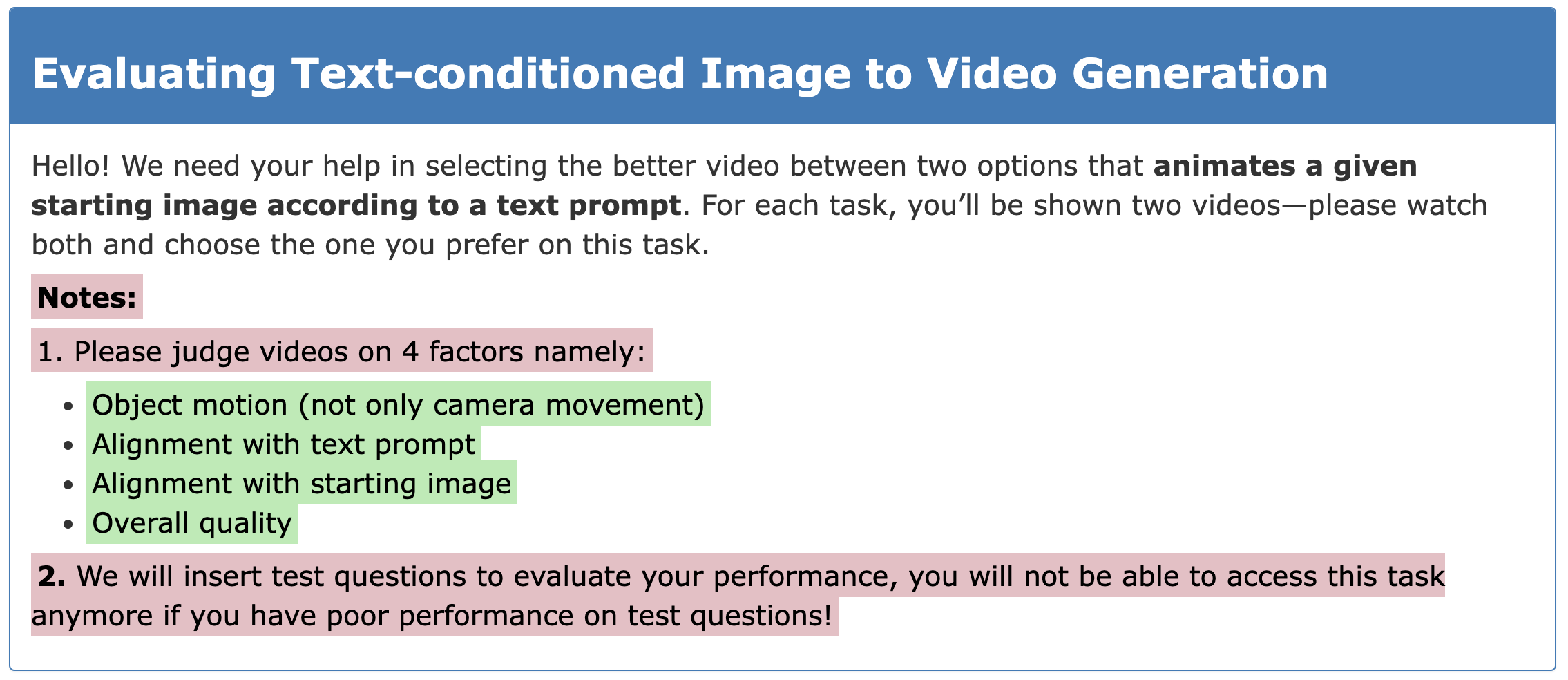}
    \includegraphics[width=0.8\linewidth]{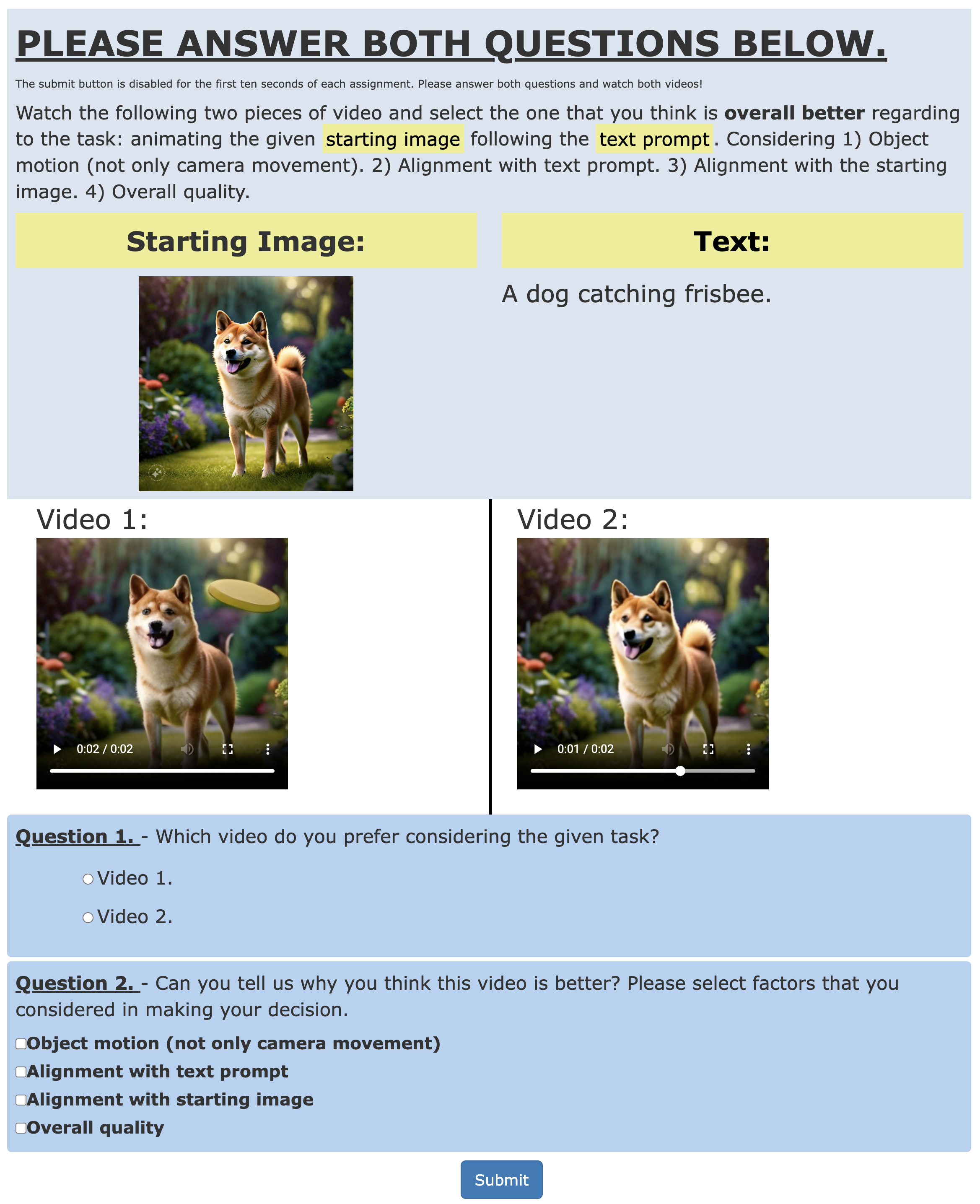}

    \captionof{figure}{\textbf{Illustration of human evaluation interface on the Amazon Mechanical Turk platform.}}
    \label{fig:mturk}
\end{figure*}

\section{Additional Implementation Details}
\label{sec:supp_details}
\subsection{Computational Costs of Optical Flow}
The average speed of optical flow generation is 2.1 videos per second on a single Nvidia A100 GPU. The speed remains the same for videos of higher resolution, as the model resizes all video frames to a fixed resolution of 960 × 520.

\subsection{Scenes in \benchmark}
\benchmark contains a total of 22 diverse scenes, designed to cover a wide range of scenarios. The detailed scenes include: \textit{car on the road, balance scale, (multiple) balloons, bird, (multiple) bulbs, butterfly, candle, child in a playground, dog, rubber duck on a pool, fish, flower, golf ball, horse, animal on a meadow, human face, human body, sun, tide, traffic light, tree, and volcano.}

\subsection{Human Evaluation}
We show the human evaluation interface in~\ref{fig:mturk}. 
During the evaluation process, the annotators are required to read the instructions first and then answer the two questions based on the specified criteria.

\end{document}